\documentclass[11pt]{article} 
\usepackage{rldm,palatino}
\usepackage{graphicx}
\usepackage{amsmath}
\usepackage{amssymb}
\usepackage{mathtools}
\usepackage{amsthm}
\usepackage{algorithm}
\usepackage{algorithmic}
\usepackage{wrapfig}

\usepackage[capitalize,noabbrev]{cleveref}

\newcommand{\TV}{D_{\mathrm{TV}}}
\newcommand{\KL}{D_{\mathrm{KL}}}

\newcommand{\piTilde}{\tilde{\pi}}
\newcommand{\TVmax}{D_{\mathrm{TV}}^{\mathrm{max}}}
\newcommand{\KLmax}{D_{\mathrm{KL}}^{\mathrm{max}}}
\newcommand{\clip}{\mathrm{clip}}

\theoremstyle{plain}
\newtheorem{theorem}{Theorem}[section]
\newtheorem{proposition}[theorem]{Proposition}

\theoremstyle{definition}
\newtheorem{definition}[theorem]{Definition}

\theoremstyle{remark}

\title{Trust-Region-Free Policy Optimization for Stochastic Policies}

\author{
Mingfei Sun \\
University of Oxford\\
\texttt{mingfei.sun@cs.ox.ac.uk} \\
\And
Benjamin Ellis \\
University of Oxford\\
\texttt{benjamin.ellis@keble.ox.ac.uk} \\
\And
Anuj Mahajan \\
University of Oxford\\
\texttt{anuj.mahajan@cs.ox.ac.uk} \\
\AND
Sam Devlin \\
Microsoft Research \\
\texttt{sam.devlin@microsoft.com} \\
\And
Katja Hofmann \\
Microsoft Research \\
\texttt{katja.hofmann@microsoft.com} \\
\And
Shimon Whiteson \\
University of Oxford\\
\texttt{shimon.whiteson@cs.ox.ac.uk} \\
}

\begin{document}

\maketitle

\begin{abstract}
   Trust Region Policy Optimization (TRPO) is an iterative method 
   that simultaneously maximizes a \emph{surrogate objective}
   and enforces a trust region constraint over consecutive policies in each iteration. 
   The combination of the surrogate objective maximization and the trust region enforcement 
   has been shown to be crucial to guarantee a monotonic policy improvement. 
   However, solving a trust-region-constrained optimization problem can be computationally intensive
   as it requires many steps of conjugate gradient and a large number of on-policy samples. 
   In this paper, we show that the trust region constraint over policies can be safely substituted
   by a \emph{trust-region-free} constraint without compromising the underlying monotonic improvement guarantee. 
   The key idea is to generalize the surrogate objective used in TRPO in a way that 
   a monotonic improvement guarantee still emerges as a result of constraining the maximum advantage-weighted ratio between policies. 
   This new constraint outlines a conservative mechanism for iterative policy optimization and
   sheds light on practical ways to optimize the generalized surrogate objective. 
   We show that the new constraint can be effectively enforced by being conservative when optimizing the generalized objective function in practice. 
   We call the resulting algorithm Trust-REgion-Free Policy Optimization (TREFree) 
   as it is free of any explicit trust region constraints. 
   Empirical results show that TREFree 
   outperforms TRPO and Proximal Policy Optimization (PPO) in terms of policy performance and sample efficiency. 
\end{abstract}


\acknowledgements{Mingfei Sun is also affiliated with Microsoft Research. 
He is partially supported by funding from Microsoft Research. 
The experiments were made possible by a generous equipment grant from NVIDIA.}

\startmain 

\section{Introduction}
Trust Region Policy Optimization (TRPO)~\cite{schulman2015trust} is an iterative method 
that optimizes stochastic policies with a trust region constraint. 
One of the key ideas in TRPO is to simultaneously optimize a \emph{surrogate objective} 
and enforce a trust region constraint over consecutive policies at each iteration. 
The use of surrogate objectives stems from the seminal work of~\cite{kakade2002approximately}, 
which modifies the policy gradient (PG) objective~\cite{sutton2000policy}
by substituting the on-policy state distribution with a distribution induced by the policy from the preceding iteration.
\cite{schulman2015trust} show that, 
despite the mismatch between what the policy update should optimize, 
i.e. the PG objective,
and what is optimized in practice, 
i.e. the surrogate objective,
a monotonic improvement guarantee for policy performance can still emerge from constraining the policy update at each iteration.
The resulting TRPO algorithm thus strictly enforces a Kullback-Leibler (KL) divergence constraint between consecutive policies,
and seeks to solve a KL-constrained surrogate objective optimization at each iteration. 
This combination of the surrogate objective maximization and the trust region enforcement 
has also been shown to be crucial for policy improvement in practice~\cite{schulman2015trust,achiam2017constrained}.

However, solving a KL-constrained optimization problem can be computationally intensive~\cite{schulman2017proximal}.
In particular, TRPO use a quadratic approximation of the KL 
that augments natural policy gradients~\cite{kakade2001natural} 
with a line-search step that critically ensures KL enforcement~\cite{schulman2015trust}. 
This procedure requires many steps of conjugate gradient
and a large number of on-policy samples making it both computationally intensive and sample inefficient~\cite{wu2017scalable}.
Many follow-up studies attempt to improve TRPO, for example by leveraging Kronecker-factored approximated curvature to approximate the trust region~\cite{wu2017scalable},
solving the KL-regularized optimization analytically via Expectation-Maximization~\cite{abdolmaleki2018maximum,hessel2021muesli},
transforming TRPO into an unconstrained optimization by policy space projections~\cite{akrour2019projections}
or integrating the constraint into differentiable layers~\cite{otto2021differentiable}.

In this paper, we propose simplifying policy optimization
by completely removing the trust region constraint, without compromising the underlying monotonic improvement guarantee. 
Specifically, instead of the framework of surrogate objective optimization~\cite{kakade2002approximately,schulman2015trust}, 
we generalize the surrogate objective used in TRPO in a way that 
a monotonic improvement guarantee still emerges as a result of constraining the maximum advantage-weighted ratio between policies. 
This new constraint is different from the trust region constraint in TRPO
in that it does not seek to impose any divergence constraint over consecutive policies. 
Instead, it outlines a conservative mechanism to bound the maximum advantage-weighted ratios in each iteration, 
and sheds light on practical ways to directly optimize the generalized surrogate objective. 
We show that the new constraint can be simply enforced 
by being conservative when optimizing the generalized objective function in practice. 
Furthermore, we present Trust-REgion-Free Policy Optimization (TREFree), 
a practical policy optimization method for optimizing stochastic policies.
Empirical results show that TREFree is effective in optimizing policies, 
outperforming TRPO and PPO in both performance and sample efficiency.

\section{Preliminaries}
\paragraph{Markov decision process (MDP).} 
Single-agent RL can be modelled as an infinite-horizon discounted Markov decision process (MDP) 
$\{\mathcal{S}, \mathcal{A}, P, r, d_0, \gamma\}$, 
where $\mathcal{S}$ is a finite set of states,
$\mathcal{A}$ is a finite set of actions,
$P: \mathcal{S} \times \mathcal{A} \times \mathcal{S} \rightarrow \mathbb{R}$ 
is the transition probability distribution, 
$r: \mathcal{S}\times\mathcal{A}\rightarrow\mathbb{R}$ is the reward function, 
$d_0:\mathcal{S}\rightarrow\mathbb{R}$ is the initial state distribution 
and $\gamma\in [0, 1)$ is the discount factor. 
Let $\pi$ denote a stochastic policy $\pi : \mathcal{S} \times \mathcal{A} \rightarrow [0, 1]$, 
the performance for a stochastic policy $\pi(a|s)$ is defined as:
$J(\pi) \triangleq \mathbb{E}_{s_0\sim d_0, a_t\sim \pi(\cdot|s_t), s_{t+1}\sim P(\cdot|s_t, a_t)}\big[ \sum_{t=0}^{\infty}\gamma^t r(s_t, a_t ) \big]$
The action-value function $Q_\pi$ 
and value function $V_\pi$ are defined as:
$Q_{\pi}(s_t, a_t) \triangleq \mathbb{E}_{t}\Big[ \sum_{l=0}^{\infty} \gamma^l r(s_{t+l}, a_{t+l}) \Big]$,
$V_{\pi}(s_t) \triangleq \mathbb{E}_{a_t\sim\pi(\cdot | s_t)}\Big[Q_{\pi}(s_t, a_t) \Big]$. 
Accordingly, the advantage function is defined as $A_{\pi}(s, a) \triangleq Q_{\pi}(s, a) - V_{\pi}(s)$.

\paragraph{TRPO.} 
Define the discounted state distribution as:
${d_{\pi}(s) \triangleq \sum_{t=0}^{\infty} \gamma^t P(s_t=s|\pi, d_0)}$.
The following equation is useful~\cite{kakade2002approximately}:
\begin{equation}\label{equ:performance-identity}
J(\piTilde) = J(\pi) + \sum_{s} d_{\piTilde}(s) \sum_{a} \piTilde(a|s)A_{\pi}(s, a). 
\end{equation}
The complex dependency of $d_{\piTilde}(s)$ on $\piTilde$ 
makes the right hand side (RHS) difficult to optimize directly. 
\cite{schulman2015trust} proposed to consider the following surrogate objective:
\begin{equation}\label{equ:trpo-surrogate}
L_{\pi}(\piTilde) \triangleq J(\pi) + \sum_{s}d_\pi(s) \sum_{a}\piTilde(a|s)A_{\pi}(s, a),
\end{equation}
where $d_{\piTilde}$ is replaced with $d_{\pi}$. 
TRPO introduces the idea of bounding the distribution change via the policy divergence.
Specifically, define $\TVmax(\pi, \piTilde)\triangleq \max_{s} \TV\big(\pi(\cdot | s), \piTilde(\cdot | s)\big)$, 
where $\TV$ is the total variation (TV) divergence. 
\begin{theorem}\label{theo:trust-region}
(\cite{schulman2015trust}) Let $\alpha \triangleq \TVmax(\pi, \piTilde)$, 
then the following bound holds:
$J(\piTilde) \geq L_{\pi}(\piTilde) - \frac{4\epsilon \gamma}{(1-\gamma)^2} \alpha^2$, 
where $\epsilon=\max_{s,a}\lvert A_\pi(s, a)\rvert$. 
\end{theorem}
Since the TV divergence and the Kullback-Leibler (KL) divergence are related as follows: 
$\TV^2(\pi, \piTilde) \leq \frac{1}{2} \KL(\pi, \piTilde)$, 
we then have the following
$J(\piTilde) \geq L_{\pi}(\piTilde) - \frac{2\epsilon \gamma}{(1-\gamma)^2} \KLmax(\pi, \tilde{\pi})$, 
where $\KLmax(\pi, \piTilde)\triangleq \max_{s}\KL(\pi, \piTilde)$. 
This forms the foundation of many policy optimization methods, 
including TRPO~\cite{schulman2015trust} 
and Proximal Policy Optimization (PPO)~\cite{schulman2017proximal}. 
The KL divergence imposed over the consecutive policies, $\pi$ and $\piTilde$, 
is also called the trust region. 
In practice, TRPO adopts a robust way to take large update steps by using a constraint
(rather than a penalty) on the KL divergence,
and also considers using the expected KL divergence, 
instead of the maximum over all states:
\begin{equation}\label{equ:trpo-objective}
\max_{\piTilde} \quad \mathbb{E}_{(s, a)\sim d_{\pi}}\Big[\frac{\piTilde(a|s)}{\pi(a|s)}A_{\pi}(s, a) \Big],
\quad \text{s.t.} \quad \mathbb{E}_{s}\big[\KL(\pi(\cdot|s), \piTilde(\cdot |s))\big] \leq \delta,
\end{equation}
where $\delta$ is a hyperparameter to specify the trust region. 
PPO with ratio clipping further simplifies such trust region constraint by leverage ratio clipping 
and considers the following optimization problem:
\begin{equation}\label{equ:ppo-objective-clip}
\max_{\tilde{\pi}} \mathbb{E}_{d_{\pi}} \big[ \min\big(\frac{\tilde{\pi}(a|s)}{\pi(a|s)} A_{\pi}, \clip(\frac{\tilde{\pi}(a|s)}{\pi(a|s)}, 1-\epsilon, 1+\epsilon)A_{\pi} \big) \big], \quad \text{ where } \epsilon \text{ is the clipping hyperparameter. }
\end{equation}

\section{Conservative policy optimization}
We show in this section that the trust region constraint over policies can be safely substituted
by a \emph{trust-region-free} constraint when we consider a generalized form of the surrogate objective function. 
We also present a monotonic improvement guarantee for stochastic policies with the generalized surrogate objective and the trust-region-free constraint. 

\subsection{Optimization of stochastic policies}
\begin{definition}\label{def:sa-function}
    Define state-action function: $A(s, a)\triangleq r(s, a) + \mathbb{E}_{s'\sim P(\cdot|s, a)}[f(s')] - f(s)$, 
    where $f$ is a function $f:\mathcal{S}\rightarrow \mathbb{R}$. 
\end{definition}
Consider updating a stochastic policy from $\pi$ to $\tilde{\pi}$ via policy gradients. 
The following proposition from~\cite{achiam2017constrained} is useful. 
\begin{proposition}\label{prop:performance-gap-stochastic}
    For any stochastic policies $\tilde{\pi}$, $\pi$, and the state-action function defined above,
    \begin{equation}\label{equ:performance-identity-generalized}
    J(\tilde{\pi}) - J(\pi) = \mathbb{E}_{s\sim d_{\tilde{\pi}}(s), a\sim\tilde{\pi}}[A(s, a)] - \mathbb{E}_{s\sim d_{\pi}(s), a\sim\pi}[A(s, a)]. 
    \end{equation}
\end{proposition}
This proposition generalizes~\eqref{equ:performance-identity} to a broader family of functions $A(s, a)$. 
One can easily verify that the advantage function of $\pi$, i.e., $A_{\pi}(s, a)$, satisfies the definition 
with $f$ function as the value function. 
In this case, \eqref{equ:performance-identity-generalized} is equivalent to~\eqref{equ:performance-identity}. 
Furthermore, this proposition implies that 
the performance difference between any two policies 
can be described by their state-action distribution shift, i.e., $d_{\tilde{\pi}}(s)\tilde{\pi}(a) - d_{\pi}(s)\pi(a)$, 
weighted by a function $A(s, a)$. 
In practice, it would be very unlikely to have access to $d_{\tilde{\pi}}(s)$. 
We thus leverage the same trick used in TRPO to substitute $d_{\tilde{\pi}}(s)$ with the state distribution induced by policy $\pi$, i.e., $d_{\pi}(s)$, 
and consider the following objective:
\begin{equation}\label{equ:new-surrogate-stochastic}
    G_\pi(\tilde{\pi}) \triangleq \mathbb{E}_{s\sim d_{\pi}(s), a\sim\pi(\cdot|s)}\big[ \big(\frac{\tilde{\pi}(a|s)}{\pi(a|s)} -1\big) A(s, a)\big]. 
\end{equation}
This new objective generalizes the surrogate objective in~\eqref{equ:trpo-objective} 
to any function defined in~\ref{def:sa-function}. 
We have the following bound, 
\begin{theorem}~\label{theo:cpo-stochastic}
    For any two stochastic policies $\tilde{\pi}$ and $\pi$, the following bound holds:
    \begin{equation*}
        J(\tilde{\pi}) - J(\pi) \geq G_{\pi}(\tilde{\pi}) - \frac{2\gamma}{1-\gamma}(\delta+\epsilon), \quad
        \text{ where } \delta = \max_{s, a} \left| \big(\frac{\tilde{\pi}(a|s)}{\pi(a|s)} -1\big) A(s, a) \right| 
        \text{ and } \epsilon = \left|\sum_{a}\pi(a|s)A(s, a) \right|. 
    \end{equation*}
\end{theorem}
To simplify further analysis, we call $\big(\frac{\tilde{\pi}(a|s)}{\pi(a|s)} -1\big)$ the \emph{ratio deviation}. 
This theorem states that the policy improvement gap can be effectively bounded by 
the maximum product of the ratio deviation $\big(\frac{\tilde{\pi}(a|s)}{\pi(a|s)} -1\big)$ 
and the state-action function $A(s, a)$. 
Moreover, as this product also appears in the definition of $G_{\pi}(\tilde{\pi})$ in~\eqref{equ:new-surrogate-stochastic}, 
one can thus consider constraining it when optimizing $G_{\pi}(\tilde{\pi})$.
This is what we call the \emph{conservative policy optimization}. 
We discuss how this conservative policy update can be implemented in practice in the next section.

Theorem~\ref{theo:cpo-stochastic} differs from Theorem~\ref{theo:trust-region} in two respects. 
First, it presents a lower bound for the performance improvement with respect to a state-action function 
defined in Definition~\ref{def:sa-function}. 
This $A(s, a)$ does not necessarily need to be the advantage function. 
Second, instead of imposing the TV constraint over the policies as in Theorem~\ref{theo:trust-region}, 
the above theorem considers the maximum product of the ratio deviation and the state-action function. 
According to~\cite{sun2022you}, the TV constraint between any two policies can be equivalently 
translated into a constraint over ratio deviations. 
In this sense, TRPO is essentially a special case of Theorem~\ref{theo:cpo-stochastic} 
by leveraging the TV to bound the ratio deviations under the assumption that  
the advantage function should be small. 
Namely, TRPO optimizes the policy regardless of how the magnitude of the advantage might change throughout optimization. 
Consequently, TRPO may fail to optimize the policy 
when the advantage function is large in magnitude at some state-action sample even though the TV divergence is well bounded at one iteration. 

\subsection{Practical policy optimization methods}

\begin{wrapfigure}{r}{0.5\textwidth}
   \vspace{-1.0em}
   \begin{minipage}{0.5\textwidth}
      \begin{algorithm}[H]
      \caption{TREFree algorithm}
      \label{algo:cpo-stochastic}
      \begin{algorithmic}
         \FOR{iterations $i=1, 2,...$}
               \FOR{actor $=1, 2,...,N$}
                  \STATE Run policy $\pi$ in environment
                  \STATE Compute advantage estimates $\hat{A}_{\pi}$
               \ENDFOR
               \FOR{epoch $=1, 2,...,K$}
                  \STATE Sample $M$ samples $\{(s, a)\}$ from previous rollouts. 
                  \STATE Compute 
                     $\mathcal{L}(\theta) \triangleq \frac{1}{M}\sum\limits_{s, a}\min\Big(\big(\frac{\piTilde_{\theta}(a|s)}{\pi(a|s)} - 1 \big) \hat{A}_{\pi}(s, a), \delta\Big)$.
                  \STATE Maximize $\mathcal{L}(\theta)$ w.r.t $\theta$ via gradient descent. 
               \ENDFOR
               \STATE $\pi \leftarrow \piTilde_{\theta}$. 
         \ENDFOR
      \end{algorithmic}
      \end{algorithm}
   \end{minipage}
   \vspace{-2.0em}
\end{wrapfigure}

We now present the practical policy optimization methods for stochastic policies. 
We first offer intuitions to understand the underlying idea of conservative policy optimization in the above theorems.

Theorem~\ref{theo:cpo-stochastic} is closely related to some existing policy optimization methods. 
For example, optimizing $G_{\pi}(\tilde{\pi})$ without any conservative constraint is equivalent to the policy gradient method~\cite{sutton2000policy}. 
Also, there are two ways to impose such conservative constraints: 
the ratio-conservative and the objective-conservative, 
which refer to removing the incentive of increasing the ratio deviations (corresponding to PPO~\cite{schulman2017proximal}) or the objective, respectively,
when optimizing the objective function.

\paragraph{Non-conservative}
With no consideration of the conservative policy update principle, 
one can directly optimize $G_{\pi}(\tilde{\pi})$ with the state-action function chosen as the advantage function of $\pi$. 
Such policy optimization is performed by \emph{policy gradient methods}~\cite{sutton2000policy}:
$\max_{\theta}\quad\mathbb{E}_{(s, a)\sim d_{\pi}}\big[\frac{\tilde{\pi}_{\theta}(s, a)}{\pi(s, a)}A_{\pi}(s, a)\big]$. 
Theorem~\ref{theo:cpo-stochastic} implies that
optimizing the above objective with the same set of sampled data for multiple times (i.e., multi-epoch optimization as in~\cite{schulman2017proximal}), 
could incur a significant degradation in policy performance,
since the product between the ratio deviation and the advantage can be large. 
Thus, applying policy gradients for multiple epoch optimization does not guarantee policy improvement~\cite{kakade2002approximately}. 

\paragraph{Ratio-conservative}
One can take into account the conservative policy update rule by constraining the ratio deviations. 
Specifically, when optimizing $G_{\pi}(\tilde{\pi})$ with the state-action function as the advantage, 
one can clip the ratio deviations to remove the incentive of inducing unexpected large deviations (i.e., \emph{ratio-conservative}), as follows
$\max_{\theta}\quad\mathbb{E}_{(s, a)\sim d_{\pi}}\big[\clip\big(\frac{\tilde{\pi}_{\theta}(s, a)}{\pi(s, a)}-1, -\lambda, \lambda\big) A_{\pi}(s, a)\big]$, 
where $\lambda>0$ is a hyper-parameter for ratio deviation clipping. 
This new objective resembles the ratio clipping objective~\eqref{equ:ppo-objective-clip} used in \emph{PPO}~\cite{schulman2017proximal}, 
which has been shown to be effective in practice, 
especially with a normalized advantage function. 
However, such ratio clipping scheme ignores the potential effect of the advantage function on the ratio deviation, 
and thus can fail to monotonically improve the policy performance 
when the advantage is large at some state-action point and dominates $\big(\frac{\tilde{\pi}_{\theta}(s, a)}{\pi(s, a)}-1\big)A_{\pi}(s, a)$ 
for a small ratio deviation. 
Furthermore, solely bounding the divergence between policies, 
as used in trust region methods for policy optimization~\cite{schulman2015trust,achiam2017constrained}, 
may not be a good option in practice, 
as it is not sufficient for policy improvement when 
the advantage function fluctuates greatly across state-action samples.

\paragraph{Objective-conservative} 
We can instead apply the clipping scheme to the objective. 
Namely, we can clip the objective directly to achieve this conservative update principle, as follows:
\begin{equation*}
\max_{\theta}\quad\mathbb{E}_{(s, a)\sim d_{\pi}}\Big[\min\Big(\big(\frac{\tilde{\pi}_{\theta}(s, a)}{\pi(s, a)}-1\big)A_{\pi}(s, a), \delta\Big)\Big], 
\end{equation*}
where $\delta>0$ is a hyper-parameter to control the conservativeness when optimizing the objective. 
The $\delta$ operates as a threshold beyond which the objective quantity have no contribution to the optimization.
We call it \emph{objective conservative}, and the resulting algorithm 
\emph{Trust-REgion-Free Policy Optimization (TREFree)} as it is free of any explicit trust region constraint. 
TREFree is detailed in Algorithm~\ref{algo:cpo-stochastic}.

\section{Experiments}
In this section, we compare TREFree with TRPO and PPO across the Mujoco continuous control tasks. 
We adopt the same strategy as in TRPO~\cite{schulman2015trust} and PPO~\cite{schulman2017proximal}
to normalize the observations, rewards, and advantages. 
Specifically, the observations, rewards and advantages are normalized to zero mean 
and unit variance using a running mean and standard deviation. 
Both observations and rewards are normalized using a running mean and standard deviation per-timestep over the whole training process,
while the advantages are normalized only within a training batch.
Moreover, we leverage the actor-critic framework by parameterizing the actor and critic 
with 2-layered perceptrons, each of which has 64 hidden units and is activated with $\tanh$. 
Also, the actor and critic share parameters by reusing the first layer of their neural networks,
which has been reported to stabilize the training and improve performance~\cite{schulman2015trust,schulman2017proximal,brockman2016openai}. 
The policy is modeled as a Gaussian distribution, with mean and variance parameterized by the actor neural network. 

\begin{figure*}
    \centering
    \includegraphics[width=1.0\linewidth]{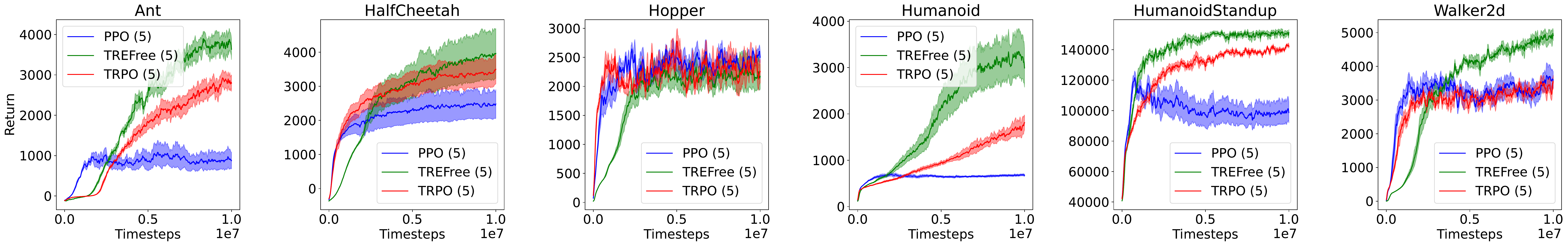}
    \caption{ Contrasting TREFree with TRPO and PPO on Mujoco benchmark tasks;
    training with a margin size $0.01$.}
    \label{fig:comparison-with-trpo-ppo-mujoco}
    \vspace{-0.5cm}
\end{figure*}

We now compare TREFree with TRPO and PPO across the Mujoco continuous control tasks with different control complexity~\cite{brockman2016openai}, 
We used the publicly available and widely used repository ({\small https://github.com/openai/baselines}) as the baseline implementation. 
For TRPO and PPO, we use the default hyper-parameters given in~\cite{schulman2015trust, schulman2017proximal}.
We also sweep over the clipping range for PPO and use the best performing value as the baseline.
Furthermore, we heuristically decay the learning rate of TREFree and PPO linearly from $0.0003$ to $0$, 
as we found this annealing strategy stabilizes training for both PPO and TREFree. 
We heuristically set $\delta$ in TREFree to $0.01$ as it is found to perform well across all tasks. 

\begin{wrapfigure}{r}{0.47\textwidth}
   \centering
   \includegraphics[width=1.0\linewidth]{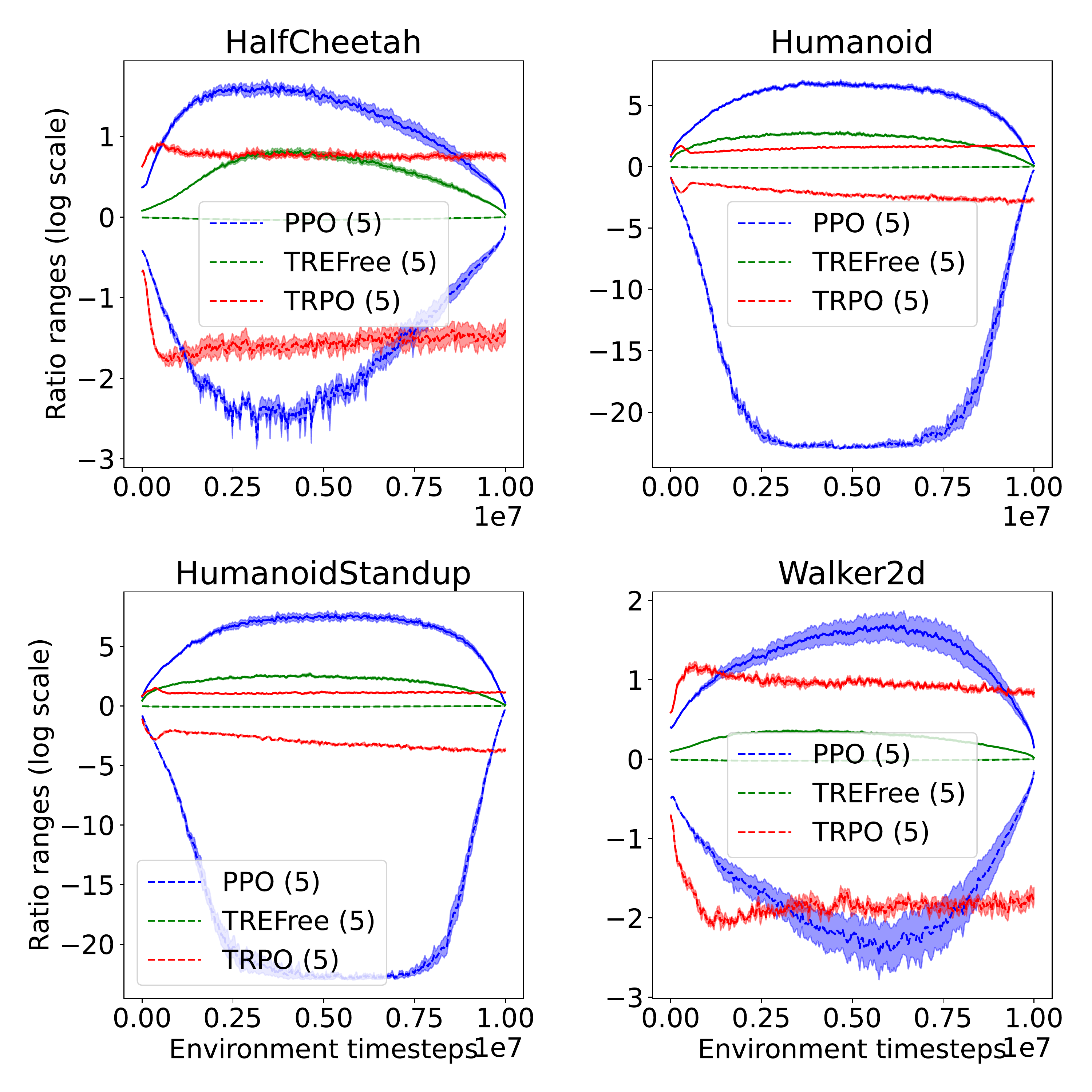}
   \caption{Contrasting ratio ranges.}
   \label{fig:ratio-analysis}
\end{wrapfigure}

The performance comparison on the Mujoco continuous control tasks is presented in Figure~\ref{fig:comparison-with-trpo-ppo-mujoco}.
Overall, TREFree performs better than the baselines on all the tasks except Hopper. 
TREFree outperforms TRPO and PPO by a large margin in terms of the final policy performance on tasks Ant, HalfCheetah, Humanoid, HumanoidStandup and Walker2d. 
These five Mujoco environments are more complicated than Hopper. 
The training curves in Figure~\ref{fig:comparison-with-trpo-ppo-mujoco} also show how TREFree often outpaces other baselines in improving policy performance. 
Though TREFree is outperformed by PPO and TRPO on Hopper, 
the performance curves of all these methods in this specific environment fluctuates greatly over time,
and overlap each other.

We also report the ratio ranges of different methods in Figure~\ref{fig:ratio-analysis}
to show the underlying differences between TREFree and the baseline methods. 
The probability ratios $\frac{\tilde{\pi}(a|s)}{\pi(a|s)}$ are an important indicator in TRPO and PPO training 
as they are closely related to the total variation divergence~\cite{schulman2017proximal}. 
Figure~\ref{fig:ratio-analysis} shows that the ratios in both TRPO and TREFree are better bounded than in PPO, where the ratios grow without bound. 
However, TREFree constrains ratios in a dramatically different way from TRPO: 
TRPO bounds ratios between $[-2, 1]$ (log-scale) in a symmetrical way,
while TREFree bounds ratios between $[0, 1]$, 
which implies that the policy is most often updated to increase the probability at empirical samples. 
This contrast in ratio ranges suggests that TREFree is fundamentally different from trust region methods. 

{\tiny \bibliography{bib}}

\bibliographystyle{plain}

\end{document}